\documentclass[10pt,twocolumn,letterpaper]{article}

\usepackage{wacv}              

\usepackage{graphicx}
\usepackage{amsmath}
\usepackage{amssymb}
\usepackage{booktabs}
\usepackage{multirow}
\usepackage{makecell}
\usepackage{enumitem}

\usepackage[accsupp]{axessibility}

\usepackage[misc]{ifsym}
\usepackage[pagebackref,breaklinks,colorlinks]{hyperref}

\usepackage[capitalize]{cleveref}
\crefname{section}{Sec.}{Secs.}
\Crefname{section}{Section}{Sections}
\Crefname{table}{Table}{Tables}
\crefname{table}{Tab.}{Tabs.}

\def\wacvPaperID{1783} 
\def\confName{WACV}
\def\confYear{2024}

\begin{document}

\title{LongFormer: Longitudinal Transformer for Alzheimer's Disease \\ Classification with Structural MRIs}



\author{Qiuhui Chen, Qiang Fu, Hao Bai, Yi Hong\textsuperscript{\Letter}\\
Department of Computer Science and Engineering, \\ Shanghai Jiao Tong University, Shanghai, 200240, China\\
{\tt yi.hong@sjtu.edu.cn}
}

\maketitle
\begin{abstract}
Structural magnetic resonance imaging (sMRI), especially longitudinal sMRI, is often used to monitor and capture disease progression during the clinical diagnosis of Alzheimer's Disease (AD).   
However, current methods neglect AD's progressive nature and have mostly relied on a single image for recognizing AD. 
In this paper, we consider the problem of leveraging the longitudinal MRIs of a subject for AD classification. To address the challenges of missing data, data demand, and subtle changes over time in learning longitudinal 3D MRIs, we propose a novel model \textbf{LongFormer}, which is a hybrid 3D CNN and transformer design to learn from image and longitudinal flow pairs. Our model can fully leverage all images in a dataset and effectively fuse spatiotemporal features for classification. We evaluate our model on three datasets, i.e., ADNI, OASIS, and AIBL, and compare it to eight baseline algorithms. Our proposed LongFormer achieves state-of-the-art performance in classifying AD and NC subjects from all three public datasets. Our source code is available online at \href{https://github.com/Qybc/LongFormer}{https://github.com/Qybc/LongFormer}.


\end{abstract}

\section{Introduction}
\begin{figure}
    \centering
    \includegraphics[width=0.5\textwidth]{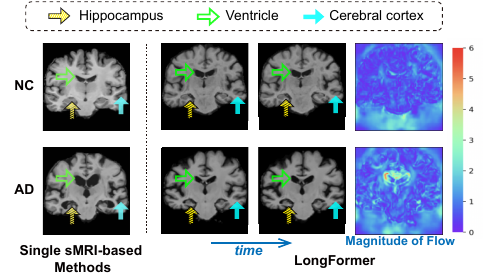}
    \caption{
    Our motivation for proposing {\it LongFormer}.
    Existing methods~\cite{zhu2021dual,li2022trans,zhang20223d,jang2022m3t} rely on the analysis of cross-sectional sMRI scans collected at a single time point, ignoring the progressive nature of AD disorder (e.g., M3T~\cite{jang2022m3t}). Our LongFormer considers the progression of brain atrophy from current and prior images, achieving the SOTA performance in AD classification.}
    \label{fig:compare}
\end{figure}

Alzheimer's Disease (AD) is one of the most common cognitive impairment diseases suffered by older people, especially in the current aging society. Medical brain scans, like Magnetic Resonance Images (MRIs), provide a non-invasive way to capture disease pathological patterns. And structural MRI (sMRI) is recommended to be a part of clinical assessment for early diagnosis of AD~\cite{qiu2022predicting}, due to its capability of characterizing brain tissue damage or loss years before the clinical symptoms appear~\cite{chan2001patterns,thompson2003dynamics}. 
As shown in the left column of Fig.~\ref{fig:compare}, compared to that of a normal control (NC) subject, the sMRI of an AD subject typically shows enlarged ventricles and hippocampus and a shrinking cerebral cortex. However, one single MRI is probably not enough to separate these two groups of subjects correctly; like the two subjects on the right column of Fig.~\ref{fig:compare}, using only one of their MRIs would lead to the wrong classification result. 
In clinical diagnosis, the degeneration speed inferred from the followed-up or longitudinal image scans is an important factor for recognizing AD subjects~\cite{saboo2021reinforcement,qiu2022predicting,hu2023vgg}. As the right column shown in Fig.~\ref{fig:compare}, by considering two scans of a subject and calculating the flow to estimate changes over time, we can separate AD subjects from normal controls more easily. 
Therefore, leveraging longitudinal brain MRIs is a promising way to study AD and help in computer-aided AD diagnosis. 

In this paper, we consider learning from 3D longitudinal structural MRIs (sMRIs) to separate AD subjects from normal controls. Researchers have collected brain MRI scans of a subject at multiple time points, resulting in a longitudinal dataset for tracking the progression of AD, e.g., the well-known ADNI dataset~\cite{jack2008alzheimer}, OASIS~\cite{marcus2007open} and AIBL~\cite{ellis2009australian}.
Most current methods proposed for MRI-based AD diagnosis treat their datasets as cross-sectional ones~\cite{jang2022m3t,zhu2021dual}, that is, they simply consider a single MRI for classification but ignore the degeneration progress included in the MRI sequence of a subject. 
However, temporal information can provide complementary supervision, solely by exploiting existing data dependency, and without requiring any additional data.
Limited work has been done to fully leverage the longitudinal dataset~\cite{cui2019rnn}. 
A recent work in~\cite{hu2023vgg} considers two time points for studying AD but only uses 2D CNN techniques, which do not fully leverage spatial information.

Learning from longitudinal 3D image volumes faces the following three challenges: (1) The missing data. Many subjects have no scans at some time points. How to handle missing data is a non-trivial task. (2) Large memory cost and limited data size. One 3D volume has millions of voxels and needs a good amount of GPU memory for a deep-learning model. At the same time, we only have hundreds of AD or NC subjects for learning. This makes it even harder to handle a sequence of 3D volumes. (3) Small longitudinal changes and subtle subject differences. A subject's changes over time are relatively small, and the differences between AD and NC subjects are subtle to recognize. 

To address these challenges, we pair each current scan of a subject with its prior image and estimate the longitudinal changes between them by computing flows via optical flow~\cite{yang2020upgrading} or deformation fields via VoxelMorph~\cite{balakrishnan2019voxelmorph}. Since we can scale the flows to normalize them, all images in the dataset can be used for learning, with no need to worry about the missing data issue. Also, because each subject can have multiple image and flow pairs, we will have thousands of input samples for learning. Besides, the pre-computed flows take some heat from the network to learn small longitudinal changes, which is experimentally demonstrated to be more effective than directly working on image pairs.   

To learn from the pair of a 3D sMRI and its flow, we develop a hybrid CNN-Transformer framework named {\it LongFormer}. Its CNN-based embedding module reduces the input size of the following transformer and the data requirement of our model. The follow-up query-based transformer adopts a deformable attention mechanism~\cite{zhu2020deformable,xia2022vision} to efficiently integrate the spatial and temporal features of the sMRI and its flow. 
Compared to eight baselines, our LongFormer achieves the state-of-the-art (SOTA) performance of classifying AD on three public datasets.

Overall, our contributions in this paper are three-fold:
\begin{itemize}[noitemsep, topsep=0pt]
    \item To our best knowledge, we are the first to explore an efficient vision Transformer on 3D longitudinal MRIs, which adaptively extracts spatiotemporal features from image and flow combinations 
    for AD classification. 
    \item We propose a novel model {\it LongFormer}, which provides a framework for learning from 4D data using 3D CNNs for embedding and query-based transformer with deformable cross-attention to fuse different sources of features flexibly and efficiently.  
    
    \item Our LongFormer achieves the SOTA performance of classifying AD and NC subjects on three public datasets, i.e., ADNI, OASIS, and AIBL. Especially, on the largest dataset ADNI, we achieve over 93\% accuracy on AD classification. 
\end{itemize}

\section{Related Work}

\noindent
\textbf{Vision Transformer.}
Transformer is firstly proposed for the sequence-to-sequence machine translation~\cite{vaswani2017attention} and currently becomes the basic component in most natural language processing tasks. Recently, the transformer has been successfully applied in computer vision, such as DETR~\cite{carion2020end} for object detection, SETR~\cite{zheng2021rethinking} for semantic segmentation, ViT~\cite{dosovitskiy2020image} and DeiT~\cite{touvron2021training} for image recognition. 
DETR proposes a new detection paradigm upon transformers, which simplifies object detection to a set prediction problem. 
Deformable DETR~\cite{zhu2020deformable} achieves better performance by using local attention and multi-scale feature maps. 
To handle videos, a sequence of image frames, SeqFormer~\cite{wu2022seqformer} adopts deformable DETR in the video instance segmentation task. 
SeqFormer proposes the query decomposition mechanism, splits instance queries at each frame, and then aggregates them to obtain a representation at the video level. 
DAT~\cite{xia2022vision} proposes a deformable attention mechanism, which is to learn a set of global keys shared among visual tokens, and can be adopted as a general backbone for vision tasks.
Our method adopts the deformable attention mechanism by selecting a set of learnable keys instead of global keys to learn visual combination representations.


\begin{figure*}[t]
    \centering
    \includegraphics[width=0.99\textwidth]{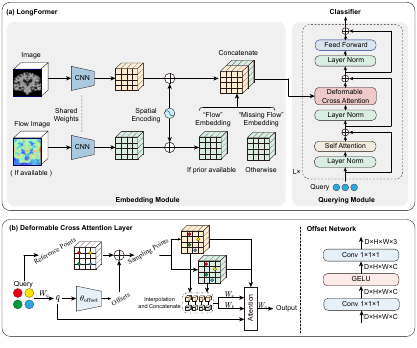}
    \caption{Illustration of our proposed LongFormer, which includes two components, i.e., the embedding module and the querying module. The deformable cross-attention layer is the core of each querying block in the querying module.}
    \label{fig:pipeline}
\end{figure*}

\noindent
\textbf{CNN-Based AD Classification.}
Deep neural networks have been widely applied for AD recognition.
Thanks to the publicly available large datasets, like 
ADNI~\cite{jack2008alzheimer}, OASIS~\cite{marcus2007open} and AIBL~\cite{ellis2009australian},
training deep models for detecting AD pathology becomes possible. In 2018, a hierarchical fully convolutional network (FCN) is proposed in~\cite{lian2018hierarchical} to learn multi-scale features from both small patches and whole brain regions to perform AD diagnosis. In 2020, a multi-modality FCN with a multilayer perceptron (MLP) model is proposed in~\cite{qiu2020development} to take both MRIs and associated subject attributes (e.g., age, gender) and is trained on ADNI and tested on multiple datasets.
In 2021, a dual-attention multi-instance deep learning model (DA-MIDL) is proposed in~\cite{zhu2021dual} to identify discriminative pathological locations for AD diagnosis using sMRIs. In 2022, a three-dimensional medical image classifier is proposed in~\cite{jang2022m3t}, using a multi-plane and multi-slice Transformer (M3T) network to classify AD using 3D MRIs. The proposed network synergically combines 3D CNN, 2D CNN, and Transformer for AD classification.
A 3D Global Fourier Network (GF-Net) is proposed in~\cite{zhang20223d} to utilize global frequency information that captures long-range dependency in the spatial domain. Trans-ResNet proposed in~\cite{li2022trans} integrates CNNs and Transformers for AD classification. 

Different from these existing works that only consider cross-sectional MRIs (i.e., a single image, ignoring time information), our model integrates longitudinal changes over time, which is an important feature for recognizing AD.

\noindent
\textbf{Longitudinal Analysis for Image-Based AD Study.}
Longitudinal study tracks the changes in brain morphology of AD subjects over time. Different from the cross-sectional setting, the longitudinal one considers multiple images from the same subject scanned at different time points as a whole for analysis. This means the dimension of the input data increases while the number of training samples decreases, which brings the risk of overfitting, especially for 3D image volumes. To address this issue, researchers use a 3D CNN to extract brain image features from each MRI, and an RNN to fuse them to extract the longitudinal changes~\cite{cui2019rnn}. 
A recent work~\cite{hu2023vgg} extracts features from image slices of longitudinal sMRIs, i.e., the baseline and its follow-up scan, and encodes them as high-level feature representation tokens by using a transformer. Since this method takes only 2D slices and two time points, it losses both spatial and temporal information of a subject's image sequence. 

After performing many experiments, we observe that directly learning from a sequence of 3D MRIs is non-trivial, due to the rapidly increased GPU memory requirement, greatly reduced input samples for learning, and the subtle longitudinal changes in the MRI sequence. Hence, we pair each image with a prior one to compute longitudinal flows between them and learn the flow for the one with the missing prior image. In this way, we can fully leverage all images in our longitudinal dataset; at the same time, we use the flow to focus on estimating the subtle longitudinal changes, which greatly reduces the learning difficulty of a network.



\section{Methodology}

Figure~\ref{fig:pipeline} presents the hybrid CNN-Transformer framework of our Longformer network, which predicts the AD classification label of a subject based on an input pair. This input pair is a combination of a current image and its associated flow which is pre-computed to measure longitudinal changes of this subject, e.g., the optical flow, or the deformation field based on image registration. 


\subsection{Pre-Computing Longitudinal Flow}
Assume a subject has an image scan at some time point, and we treat it as the current scan $\mathbf{x}^{\mathrm{curr}}$. To consider the longitudinal changes, e.g., within one year, we take the prior image one year ago and will face two situations, i.e., the prior image exists and we have the longitudinal pair that includes multiple images, $(\mathbf{x}^{\mathrm{curr}}, \mathbf{x}^{\mathrm{prior}}) \in \mathcal{D}_m$, or it does not exist and the pair includes a single image, $(\mathbf{x}^{\mathrm{curr}}, \varnothing) \in \mathcal{D}_s$. In this way, we reformulate our longitudinal dataset and construct a new one $\mathcal{D}=\mathcal{D}_{m} \cup \mathcal{D}_{s}$. 

For the image pair in the subset $\mathcal{D}_m$, we directly compute their optical flow based on~\cite{yang2020upgrading} or deformation fields using a pre-trained VoxelMorph~\cite{balakrishnan2019voxelmorph}. For the image pair in the subset $\mathcal{D}_{s}$, we also have two cases, i.e., one is the baseline image that has no prior image, and the other is a follow-up image that has a prior image within half year or 1.5 years ago. For the baseline image case, we leave the flow empty, which will be learned later in our LongFormer. For the follow-up image case, we compute the flow using its combination with the prior image and scale the flow using the age difference, as below: 
\begin{equation}
    I^{\mathrm{flow}} = \frac{\Phi(I^{\mathrm{curr}},I^{\mathrm{prior}})}{t^{\mathrm{curr}}-t^{\mathrm{prior}}} ,
\end{equation}
where $\Phi$ indicates the computation function of optical flow or deformation fields.

\subsection{LongFormer Framework}
\label{sec:longformer}
As shown in Fig~\ref{fig:pipeline}(a), our LongFormer network includes two main modules, i.e., an embedding module that extracts image and flow features based on a CNN and a querying module that fuses and learns spatiotemporal representation for AD classification based on a transformer. 

The embedding module takes the input image and flow pair $(I^\mathrm{curr}, I^\mathrm{flow})$, where $I^\mathrm{curr}\in \mathbb{R}^{D \times H \times W}$ is the 3D current image and $I^\mathrm{flow}\in \mathbb{R}^{3 \times D \times H \times W}$ is the associated flow in the vector form. 
The embedding module produces the support features $F_S \in \mathbb{R}^{C_S \times D_S \times H_S \times W_S}$ for the follow-up querying module. The querying module is then responsible for learning query representation, that is, the query features $F_Q \in \mathbb{R}^{C_Q \times N_Q}$ based on the learnable query $Q$ and the support features $F_S$. Finally, the classification head produces a prediction based on the learned query features $F_Q$. Overall, our LongFormer can be briefly expressed as
\begin{equation}
    \begin{aligned}
    F_{S} & =\operatorname{Embedding}(I^\mathrm{curr}, I^\mathrm{flow}), \\
    F_{Q} & =\operatorname{Querying}\left(F_{S}, Q\right), \\
    O & =\operatorname{Classification \, Head}\left(F_{Q}\right),
    \end{aligned}
\end{equation}
where $O$ indicates the final prediction output.

\noindent
\textbf{(1) Embedding Module.}
Since the follow-up querying module will further extract spatiotemporal features for classification, this embedding module adopts a 3D CNN network, e.g., a ResNet~\cite{he2016deep}, DenseNet~\cite{huang2017densely}, which serves as a backbone network to provide visual embeddings of individual images and flows. This CNN backbone is a reasonable choice because of the CNN’s inductive biases~\cite{d2021convit,park2022vision}, and it helps in efficiency by reducing the input size of the following transformer-based querying module.  

To further reduce the number of parameters of the backbone network, we prefer the weight-sharing technique for the two branches of the embedding module. However, the input image and flow have different sizes; therefore, we first apply a convolutional layer before sharing weights, i.e.,
\begin{equation}
    F_{S} = \operatorname{Embedding}\left(f_{a_{1}}(I^\mathrm{curr}), f_{a_{2}}(I^\mathrm{flow})\right).
\end{equation}
Here, $f_{a_{1}}$ and $f_{a_{2}}$ are convolutional layers taking one and three input channels, respectively. 


To extract support features $F_S$, we take the feature maps at the stage $S5$ of the backbone network, where the spatial resolution is $1/2^5=1/32$ of the input image. While the number of channels of the features is $C = 1024$. 
When no prior image is available, the flow embedding branch is not used but replaced by a learnable embedding, which is replicated across the spatial dimensions.


\noindent
\textbf{(2) Querying Module.} 
The support feature $F_S$ generated by the embedding module is a concatenation of features for a current image and its associated longitudinal flow. To fuse these two sets of feature maps, the querying module utilizes a query-based transformer as in Fig~\ref{fig:pipeline}, which captures patch embedding interactions and aggregates them to learn a fixed-length token representation. 

As shown in Fig~\ref{fig:pipeline}(a), our querying module is a stack of $L$ querying blocks. Each block considers the support features $F_S$ from the embedding module with spatial position encoding and the query features $F_Q$ from the previous layer. Using the self- and cross-attention operators, the querying block gradually learns and refines the query features. Also, to improve the flexibility of receptive fields in transformer layers, we adopt a deformable cross-attention design similar to~\cite{xia2022vision} (see Fig.~\ref{fig:pipeline}(b)), where we replace \textbf{q} to learnable queries, instead of image features itself. This transformer mechanism equipped by our querying design satisfies the need of generating flexible query features $F_Q$, which has large receptive fields and strong representation ability, as demonstrated by our experimental results. Details on querying blocks are included in Sec.~\ref{sec:querying_block}. 

\noindent
\textbf{(3) Classification Head.}
To perform classification, we can use the first query to produce a representative feature for the classifier. Another choice is using multiple uniformly distributed queries indicating different views of the subject to vote for the final prediction. In this paper, we intend to use the first one to query the category. For the loss function, since we work on the binary classification (i.e., AD and NC), we simply use the binary cross-entropy loss.

Next, let us discuss in detail the building block of our LongFormer, i.e., the querying blocks.

\subsection{LongFormer's Querying Block}
\label{sec:querying_block}
Each querying block takes two inputs. Take the $l$-th querying block for example. This block inputs the query features $F_Q^{l-1}$ from the last layer and the support features $F_S$ from the embedding module, and then outputs $F_Q^{l}$ for the next layer. For the first querying block, we initialize $F_Q^{0}$ randomly. All $L$ querying blocks update the query features at each iteration. In our experiments, we set $L$ to be 6. 

Each querying block consists of three layers, i.e., a self-attention layer, a deformable cross-attention layer, and a feed-forward layer (FFN). We formulate them as 
\begin{equation}
    \begin{aligned}
    \hat{F}_{Q}^{l} & =\operatorname{Self-Attention}\left(\operatorname{LN}(F_{Q}^{l-1})\right)+F_{Q}^{l-1}, \\
    \hat{F}_{Q}^{l} & =\operatorname{Deformable \ Cross-Attention}\left(\operatorname{LN}(\hat{F}_{Q}^{l}),F_S\right)+\hat{F}_{Q}^{l}, \\
    F_{Q}^{l} & =\operatorname{FFN}\left(\operatorname{LN}(\hat{F}_{Q}^{l})\right)+\hat{F}_{Q}^{l} .
    \end{aligned}
\end{equation}
Here, $\operatorname{LN}$ is a layer normalization~\cite{ba2016layer} to normalize features before each attention and FFN module. The self-attention layer is a classical \textbf{qkv}-based multi-head self-attention~\cite{vaswani2017attention}, where \textbf{q}, \textbf{k}, and \textbf{v} are all from the learnable query features $F_{Q}^{l-1}$. The deformable cross-attention layer is a \textbf{qkv}-based multi-head cross-attention, where \textbf{q} is from the query features $\hat{F}_{Q}^{l}$, while \textbf{k} and \textbf{v} are from the support features $F_{S}$.

\noindent
\textbf{Deformable Cross-Attention Layer.} This layer is responsible for integrating features from images and flows, that is, it fuses and extracts spatiotemporal features for AD classification. 
As illustrated in Fig.~\ref{fig:pipeline}(b), this deformable attention layer computes cross-attention at multiple sample points $P_Q \in \mathbb{R}^{N \times 3}$, which have flexible locations within the supported features $F_S \in \mathbb{R}^{C_S \times D_S \times H_S \times W_S}$. Here, $N < D_S \times H_S \times W_S$ is the number of target positions, which can be divided into a uniform grid with $N=D_G\times H_G\times W_G$ points and are treated as the references. In particular, these reference points are uniformly located in the 3D coordinates $[(0, 0, 0), (D_G-1, H_G-1, W_G-1)]$. 
Then, we normalize them into the range $[-1, +1]$ according to the grid shape, where $(-1, -1, -1)$ indicates the top-left corner and $(+1, +1, +1)$ indicates the bottom-right corner. 

To estimate the offset for each reference point, the learnable queries, i.e., $F_Q \in \mathbb{R}^{N \times d}$ (d denotes the number of channels), are projected linearly to the query tokens $q = F_Q W_q$, and then fed into a lightweight sub-network $\theta$ offset(·) to generate the offsets $\Delta p = \theta_{\text {offset}}(q)$. To stabilize the training process, we scale the amplitude of $\Delta p$ by a pre-defined factor $s=2$ to prevent those large offsets, i.e., $\Delta p \leftarrow s \tanh (\Delta p)$. Then the features are sampled at the locations of deformed points as keys and values, followed by a set of linear projections:
\begin{equation}
    \begin{array}{l}
    q=F_Q W_{q}, \; \tilde{k}=\tilde{F_S} W_{k}, \;  \tilde{v}=\tilde{F_S} W_{v} \\
    \text { with } \tilde{F_S}=\phi(F_S ; P_Q+\Delta p),  \; \Delta p=\theta_{\text {offset }}(q), 
    \end{array}
\end{equation}
where $W_q, W_k, W_v$ are projection matrices, $\tilde{k}$ and $\tilde{v}$ represent deformed key and value embeddings, respectively, and $\phi(\cdot;\cdot)$ is a  sampling function using trilinear interpolation.

\begin{table*}[t]
    \centering
    \begin{tabular}{lcccccc}
        \toprule
        \multirow{2}{*}{Method} & \multicolumn{2}{c}{ADNI} & \multicolumn{2}{c}{OASIS}  & \multicolumn{2}{c}{AIBL} \\
        \cmidrule(r){2-3} \cmidrule(r){4-5} \cmidrule(r){6-7}
        
        & Accuracy & AUC & Accuracy & AUC  & Accuracy & AUC \\
        \midrule
        3D ResNet50 ~\cite{he2016deep} & 82.72\% & 81.04\% & 66.15\% & 65.94\%  & 72.99\% & 51.04\%      \\
        3D ResNet101 ~\cite{he2016deep} & 85.19\% & 83.57\% & 69.23\% & 69.02\% & 76.43\% & 57.94\%      \\
        3D ResNet152 ~\cite{he2016deep} & 87.65\% & 86.92\% & 70.77\% & 70.49\% & 77.59\% & 61.95\%   \\
        3D DenseNet121 ~\cite{iandola2014densenet} & 88.89\% & 87.98\% & 72.31\% & 72.68\% & 82.76\% & 73.26\%   \\
        3D ViT~\cite{dosovitskiy2020image} & 80.24\% & 81.35\% & 67.69\% & 67.13\% & 73.56\% & 76.60\% \\
        
        MRNet~\cite{bien2018deep} & 87.96\% & 93.16\% & 70.77\% & 81.97\% & 75.86\% & 75.17\% \\ 
        MedicalNet~\cite{chen2019med3d} & 88.89\% & 88.80\% & 73.85\%  & 72.72\% & 82.76\% & 79.07\% \\
        M3T~\cite{jang2022m3t} & 90.05\% & 88.78\% & 80.47\% & 81.67\% & 82.35\% & 80.26\% \\
        \midrule
        Single Image & 90.67\% & 89.03\% & 80.23\% & 80.71\% & 82.78\% & 78.42\% \\
        w/ Prior Image (see Fig.~\ref{fig:prior_image}) & 91.42\% & 91.37\% & 81.38\% & \textcolor{blue}{82.14\%} & 83.94\% & 84.14\%\\
        LongFormer w/ VoxelMorph (ours) & \textcolor{blue}{92.70\%} & \textbf{93.75\%}  & \textcolor{blue}{81.46\%} & 81.50\% & \textbf{85.77\%} & \textbf{84.47\%}\\
        LongFormer w/ Optical Flow (ours) & \textbf{93.43\%} & \textcolor{blue}{93.30\%} & \textbf{82.35\%} & \textbf{82.86\%} & \textcolor{blue}{84.09\%} & \textcolor{blue}{84.24\%} \\
        \bottomrule
    \end{tabular} 
    \caption{Quantitative results of our LongFormer and baseline methods on classifying AD and NC subjects from three datasets. The top eight baselines are existing methods and the flowing four methods are variants of our Longformer for ablation study. The best results are in \textbf{bold} and the second best ones are colored in \textcolor{blue}{blue}. 
    }
    \label{tab:comparison1}
\end{table*}

\begin{figure*}[t]
    \centering
    \includegraphics[width=0.8\textwidth]{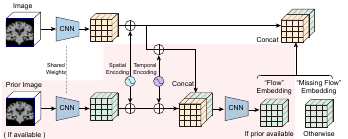}
    \caption{Ablation Study: Modifying the embedding module of our LongFormer to take a prior image instead of a pre-computed longitudinal flow. This architecture is similar to~\cite{bannur2023learning}.}
    \label{fig:prior_image}
\end{figure*}

        

\begin{figure*}[t]
    \centering
    \includegraphics[width=\textwidth]{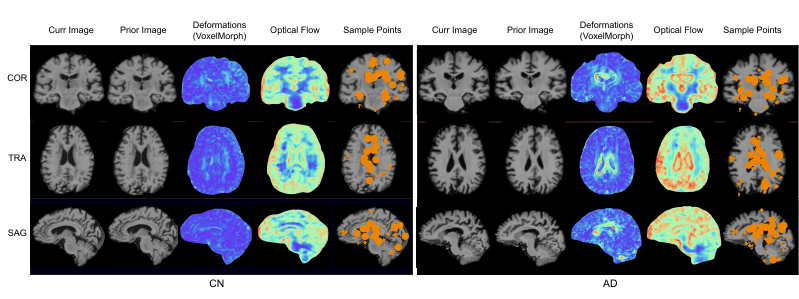}
    \caption{Sagittal (SAG), transverse (TRA), and coronal (COR) view visualization of longitudinal flows and deformable attention of our LongFormer, for AD and NC subjects sampled from the ADNI test set. 
    The flows are visualized using their magnitudes. 
    We use the jet colormap in that the red color is close to one (indicating high activated value) and the blue close to zero (indicating low activated value). The orange circles show the sample points with the highest propagated attention scores at multiple heads. A larger radius indicates a higher score. (Best viewed in color)}
    \label{fig:visualization}
\end{figure*}

\section{Experiments}

\subsection{Experimental Datasets}
We evaluate our model on the following three datasets.

\noindent
\textbf{ADNI~\cite{jack2008alzheimer}.} The Alzheimer’s Disease Neuroimaging Initiative (ADNI) dataset includes 5265 1.5T/3T T1-weighted structural MRI (sMRI) scans collected from 1306 subjects with visits at one or multiple time points across four ADNI phases (i.e., ADNI-1, ADNI-2, ADNI-GO, and ADNI-3).
These subjects are divided into two categories, i.e., AD (including 413 subjects) and NC (including 893 subjects), according to the standard clinical criteria, e.g., the Mini-Mental State Examination (MMSE) scores and the Clinical Dementia Rating (CDR).
The original structural MRI data downloaded from the ADNI website went through a series of pre-processing steps, including denoising, bias field correction, skull stripping, and affine registration to the SRI24 atlas. Then, we resample the image volumes, resulting in images of size $224 \times 224 \times 224$, with a resolution of $1.75mm \times 1.75mm \times 1.75mm$. 

\noindent
\textbf{OASIS~\cite{marcus2007open}.}
Open Access Series of Imaging Studies (OASIS) is a project aimed at making neuroimaging data sets of the brain freely available to the scientific community. We use OASIS 2 of this dataset, which has longitudinal data for evaluation. This dataset consists of 335 T1-weighted sMRI scans collected from 135 subjects, including both AD subjects and healthy volunteers. 
We have a similar pre-processing step to ADNI. 
As a result, we also have images of size $224 \times 224 \times 224$ with a resolution of $1.75mm \times 1.75mm \times 1.75mm$. 

\noindent
\textbf{AIBL~\cite{ellis2009australian}.}
The Australian Imaging, Biomarker \& Lifestyle flagship study of aging (AIBL) is a study to discover biomarkers, cognitive characteristics, and health and lifestyle factors that determine the subsequent development of symptomatic Alzheimer’s Disease (AD). 
This dataset includes 997 T1-weighted structural MRI (sMRI) scans collected from 456 subjects with Alzheimer’s disease (AD) and healthy volunteers.
We separate training and evaluation sets just like the ADNI dataset, and use the raw image data downloaded from the official website without any pre-processing steps, except for a simple alignment with centering the brain region and normalizing the image intensity. 
Then, we resample the image volumes, resulting in images of size
$224 \times 224 \times 224$, with a resolution of $1.6mm \times 0.9mm \times 0.9mm$. 

For all images, we normalize the image intensity to zero mean and unit variance. For all datasets, 
we {\it subject-wisely} divide them into 80\% for training and 20\% for testing. 

\subsection{Experimental Settings}

We adopt a 3D DenseNet121~\cite{iandola2014densenet} as our CNN backbone to extract visual features from input pairs. The 3D CNN inputs 3D volumes or vector fields of size $224\times224\times224$ and summarizes them into 3D representation features of size $7\times7\times7$ with 1024 channels. In the transformer part, we use $L=6$ querying blocks and 125 learnable queries. The deformable cross-attention layer has a hidden dimension of 512 and 8 attention heads.

We use the Adam optimizer with $\beta_1$ = 0.9 and $\beta_2$ = 0.999 for 100 epochs with a learning rate of 5e-5, and the batch size is 8. Our model is implemented using PyTorch-1.12 and is trained on four NVIDIA GeForce RTX 3090 GPUs. 
To evaluate the classification performance, we use the classification accuracy (ACC) and the area under receiver operating characteristic curve (AUC) as our evaluation metrics. 

\subsection{Comparison Results}
We compare our LongFormer with conventional 3D classification methods based on 3D ResNet (50, 101, 152)~\cite{he2016deep}, 3D DenseNet121~\cite{huang2017densely}, and ViT~\cite{dosovitskiy2020image}, since these methods have been widely used for AD classification~\cite{zhang20213d,yang2018visual,ruiz20203d,li2021comparison,korolev2017residual,karasawa2018deep,ebrahimi2020introducing}. 
We implement the 3D ViT, which is composed of pure-transformer networks~\cite{dosovitskiy2020image}, and M3T~\cite{jang2022m3t} which is composed of 2D CNN, 3D CNN, and transformer networks. 
In the M3T model, the sequence in the transformer is applied to extracted 3D patch embedding with a size of $16 \times 16 \times 16$, and the projection dimension is 512. We also compare our model with MRNet~\cite{bien2018deep} and MedicalNet~\cite{chen2019med3d}. 
The MRNet used in our experiment is based on 2D ResNet50 because it has better performance than AlexNet. MedicalNet is pre-trained on 23 medical databases.

Table~\ref{tab:comparison1} reports the quantitative results for comparing our method with baselines. Overall, our LongFormer performs best among these methods on all datasets. Even only considering a single image, our CNN-Transform hybrid design achieves comparable or better performance than the current SOTA method, M3T. Compared to 3D ViT, our method has a large performance gain, over 10\% improvement on average, which indicates the effectiveness of using CNN as the backbone. We argue that a pure transformer needs a large amount of data for training, while our hybrid design alleviates this data demand greatly, which suits our datasets. 

Figure~\ref{fig:visualization} visualizes the longitudinal flows generated by VoxelMorph and optical flow, respectively, and the responses of our LongFormer on classifying AD and CN subjects. Since our LongFormer with VoxelMorph and optical flow produce similar responses, we only present the one using VoxelMorph. This response map visualizes those sample points in the deformable cross-attention layer, which are located at those important regions for recognizing AD and support our hypothesis in Fig.~\ref{fig:compare} at the beginning.

\begin{table*}[t]
    \centering
    \begin{tabular}{lcccccc}
        \toprule
        \multirow{2}{*}{Backbone} & \multicolumn{2}{c}{ADNI} & \multicolumn{2}{c}{OASIS}  & \multicolumn{2}{c}{AIBL} \\
        \cmidrule(r){2-3} \cmidrule(r){4-5} \cmidrule(r){6-7}
        
        & Accuracy (\%) & AUC (\%) & Accuracy (\%) & AUC (\%)  & Accuracy (\%) & AUC (\%) \\
        \midrule
        3D ResNet50 & 87.33 \small{\color{red}{(+4.61)}} & 85.41 \small{\color{red}{(+4.37)}} & 70.59 \small{\color{red}{(+4.44)}} & 73.33 \small{\color{red}{(+7.39)}} & 70.45 \small{\color{red}{(+2.54)}} & 72.62 \small{\color{red}{(+21.58)}}  \\
        3D ResNet101 & 86.50 \small{\color{red}{(+1.31)}} & 86.23 \small{\color{red}{(+2.66)}} & 76.47 \small{\color{red}{(+7.24)}} & 71.43 \small{\color{red}{(+2.41)}} & 77.27 \small{\color{red}{(+0.84)}} & 79.40 \small{\color{red}{(+21.46)}} \\
        3D ResNet152 & 92.67 \small{\color{red}{(+5.02)}} & 92.64 \small{\color{red}{(+5.72)}} & 79.94 \small{\color{red}{(+9.17)}} & 80.71 \small{\color{red}{(+10.22)}} & 81.82 \small{\color{red}{(+4.23)}} & 80.40 \small{\color{red}{(+18.45)}} \\
        3D DenseNet121 & \textbf{93.43} \small{\color{red}{(+4.54)}} & \textbf{93.30} \small{\color{red}{(+5.32)}} & \textbf{82.35} \small{\color{red}{(+10.04)}} & \textbf{82.86} \small{\color{red}{(+10.18)}} & \textbf{84.09} \small{\color{red}{(+1.33)}} & \textbf{84.24} \small{\color{red}{(+10.98)}}\\
        \bottomrule
    \end{tabular} 
    \caption{Classification performance comparison using different 3D CNN backbones in our LongFormer with optical flow. The numbers in \textcolor{red}{red} indicate the improved performance, compared to using the backbone network for classification directly, as reported in Table~\ref{tab:comparison1}.
    }
    \label{tab:comparison3}
\end{table*}

\begin{table*}[t]
    \centering
    \begin{tabular}{ccccccc}
        \toprule
        \multirow{2}{*}{\#Queries} & \multicolumn{2}{c}{ADNI} & \multicolumn{2}{c}{OASIS}  & \multicolumn{2}{c}{AIBL} \\
        \cmidrule(r){2-3} \cmidrule(r){4-5} \cmidrule(r){6-7}
        
        & Accuracy & AUC & Accuracy & AUC  & Accuracy & AUC \\
        \midrule
        27 & 89.40\% & 88.72\% & 80.46\% & \textcolor{blue}{81.99\%} & 83.91\% & 79.22\%  \\
        64 & 92.59\% & 91.58\% & 81.61\% & 81.67\% & \textbf{84.48\%} & 80.26\% \\
        125 & \textcolor{blue}{93.43\%} & \textbf{93.30\%} & \textbf{82.35\%} & \textbf{82.86\%} & \textcolor{blue}{84.09\%} & \textbf{84.24\%} \\
        343  & \textbf{93.82\%} & \textcolor{blue}{92.64\%} & \textbf{82.35\%} & 80.71\% & 82.76\% & \textcolor{blue}{82.09\%} \\
        \bottomrule
    \end{tabular}
    \caption{Classification performance comparison using different numbers of queries in our LongFormer with optical flow. The best results are in \textbf{bold} and the second best ones are colored in \textcolor{blue}{blue}. }
    \label{tab:query_num}
\end{table*}

\subsection{Ablation Study}
\noindent
\textbf{(1) Longitudinal Flow.} Firstly, we demonstrate the necessity of using the pre-computed longitudinal flow and the way of computing it. Therefore, we consider four cases: (1) using a single image, that is, whether we need longitudinal images for AD classification, or a cross-sectional setting is enough; (2) using the prior image directly, that is, whether the network can figure out the longitudinal changes from the image sequence directly; (3) using deformation fields computed by VoxelMorph, which is one way to compute the longitudinal flow; and (4) using optical flow to compute the flow, an alternative. 

Specifically, for the second case, we modify our LongFormer network and implement a version of learning from image sequences directly, according to~\cite{bannur2023learning}. In this variant, after the spatial position encoding, we add a temporal encoding $\textbf{T} \in \mathbb{R}^{T \times C_S}$, which takes into account the time information (see Figure ~\ref{fig:prior_image}). This modified embedding module outputs the concatenation of two features, which is an `aggregated' representation of longitudinal flow features anchored on the current and prior image features.


As reported in Table~\ref{tab:comparison1}, experiments with longitudinal settings outperform the one with a single image, which indicates longitudinal information helps in AD classification. 
Compared with directly working on image sequences, our model performs better when taking pre-computed longitudinal flows for learning. The two methods for computing the longitudinal flows, e.g., VoxelMorph and optical flow, perform equally well on our datasets.  



\noindent
\textbf{(2) Image Embedding.} Next, we experiment with the effect of our CNN backbone on AD classification. In this ablation study, we choose LongFormer with optical flow and compare four different backbone networks. As shown in Table~\ref{tab:comparison3}, a deeper ResNet provides a better image embedding, while our choice, i.e., DenseNet121, performs best among these four backbones. This experiment demonstrates the backbone network plays an essential role in our LongFormer. Also, compared to directly using these backbones for classification as reported in Table~\ref{tab:comparison1}, our LongFormer with transformer further improves the classification performance in all cases. This indicates the effectiveness of using a transformer to integrate longitudinal changes.


\noindent
\textbf{(3) Learnable Queries.} Lastly, we would like to explore an optimal number of learnable queries for our LongFormer. As reported in Table~\ref{tab:query_num}, we test on four different numbers of queries, and the one having 125 queries performs the best for most cases, which is set as default for other experiments.

\section{Conclusion and Discussion}
In this paper, we have proposed an effective CNN-Transformer architecture, {\it LongFormer}, for Alzheimer's disease classification based on longitudinal sMRI volumes. LongFormer adopts attention mechanisms with learnable queries and deformable cross-attention to integrate both spatial and temporal information in an image scan and its longitudinal flow. 
Our proposed method provides a way to address the issues of missing data, the limited size of a dataset, and the subtle changes over time and subject differences in AD classification based on 3D longitudinal MRIs. The ablation studies demonstrate the effectiveness of our model design. Compared to multiple recent baselines, our model achieves the SOTA AD classification performance on three public datasets. 

\noindent
\textbf{Limitations and Future Work.} Currently, we only consider two scans of a subject, i.e., a current image and its prior one, to compute the longitudinal flow. To have a more accurate estimation of the flow, we could apply image regression~\cite{hong2017fast,ding2019fast} on an image sequence. Further improving the quality of the estimated longitudinal flow would help improve the classification performance of our LongFormer.  Also, our experiments focus on binary classification, which can be straightforwardly extended to multi-label classification, e.g., classifying AD, Mild Cognitive Impairment (MCI), and NC. 
Besides, our LongFormer only takes MRI scans for AD classification; however, other attributes, like age, gender, lab results, and other image modalities, like PET, fMRI, etc., are all beneficial for AD diagnosis. How to integrate them in a uniform framework for our task is an interesting research topic and left as our future work.

\section*{Acknowledgments}
This work was supported by NSFC 62203303 and Shanghai Municipal Science and Technology Major Project 2021SHZDZX0102.

{\small
\bibliographystyle{ieee_fullname}
\bibliography{egbib}
}

\end{document}